\documentclass[10pt]{article} 
\usepackage[preprint]{rlc}

\usepackage{amssymb}            
\usepackage{mathtools}          
\usepackage{mathrsfs}           
\mathtoolsset{showonlyrefs}     
\usepackage{graphicx}           
\usepackage{subcaption}         
\usepackage[space]{grffile}     
\usepackage{url}                
\usepackage{algorithm}
\usepackage{algpseudocode}

\usepackage{paralist}
\usepackage{tikz}
\usetikzlibrary{positioning, shapes, arrows, backgrounds}
\usepackage{siunitx}
\usepackage{layouts}
\usepackage{bbm}
\usepackage{accents}
\usepackage{amsfonts}
\usepackage{wrapfig}

\DeclarePairedDelimiter\abs{\lvert}{\rvert}

\newcommand{\fakepar}[1]{\vspace{1mm}\noindent\textbf{#1.}}

\newcommand*\samethanks[1][\value{footnote}]{\footnotemark[#1]}

\title{Contextualized Hybrid Ensemble Q-learning: \\ Learning Fast with Control Priors}

\author{Emma Cramer\thanks{These authors contributed equally.}  \\
    emma.cramer@dsme.rwth-aachen.de \\
    Institute for Data Science in Mechanical Engineering\\
    RWTH Aachen University
    \And
    Bernd Frauenknecht\samethanks  \\
    bernd.frauenknecht@dsme.rwth-aachen.de \\
    Institute for Data Science in Mechanical Engineering\\
    RWTH Aachen University
    \And
    Ramil Sabirov\samethanks  \\
    ramil.sabirov@dsme.rwth-aachen.de \\
    Institute for Data Science in Mechanical Engineering\\
    RWTH Aachen University
    \And
    Sebastian Trimpe \\
    trimpe@dsme.rwth-aachen.de \\
    Institute for Data Science in Mechanical Engineering\\
    RWTH Aachen University}

\begin{document}

\maketitle

\begin{abstract}
Combining Reinforcement Learning (RL) with a prior controller can yield the best out of two worlds: RL can solve complex nonlinear problems, while the control prior ensures safer exploration and speeds up training.
Prior work largely blends both components with a fixed weight, neglecting that the RL agent's performance varies with the training progress and across regions in the state space. 
Therefore, we advocate for an adaptive strategy that dynamically adjusts the weighting based on the RL agent's current capabilities. 
We propose a new adaptive hybrid RL algorithm, \emph{Contextualized Hybrid Ensemble Q-learning} (CHEQ). 
CHEQ combines three key ingredients: (i) a time-invariant formulation of the adaptive hybrid RL problem treating the adaptive weight as a context variable, (ii) a weight adaption mechanism based on the parametric uncertainty of a critic ensemble, and (iii) ensemble-based acceleration for data-efficient RL.
Evaluating CHEQ on a car racing task reveals substantially stronger data efficiency, exploration safety, and transferability to unknown scenarios than state-of-the-art adaptive hybrid RL methods.
\end{abstract}
%
%
%
%
%

\section{Introduction}
\label{sec:introduction}
Deep reinforcement learning (RL) methods have shown great success in challenging control problems such as gameplay ~\citep{mnih_human-level_2015, silver_general_2018, openai_dota_2019} and robotic manipulation ~\citep{gupta_reset-free_2021, buchler_learning_2022}. Despite the great potential of RL methods, their data inefficiency, unstructured exploration behavior, and inability to generalize to unknown scenarios represent a significant hurdle to their application to real-world problems.

A prime reason for limited real-world applications is the task-agnostic architecture of state-of-the-art RL approaches~\citep{schulman_2017, haarnoja_soft_2018} that does not incorporate prior knowledge on how to solve the task at hand. 
In contrast, control theory provides a rich set of methods for deriving near-optimal controllers in many applications.
This motivates the drive for hybrid RL methods ~\citep{silver_residual_2018, johannink_residual_2019} that blend control priors with deep RL policies. Hybrid algorithms thus combine the prior controller's generalization capabilities and informed behavior with the power of deep RL for solving general nonlinear problems.

Notwithstanding the conceptual benefit of hybrid RL formulations, how to systematically combine the control prior with the RL agent largely remains an open problem. The majority of prior work~\citep{silver_residual_2018, johannink_residual_2019, schoettler_deep_2020, ceola_resprect_2024} proposes a fixed weighting between the control prior and the RL agent. A fixed blending, however, disregards the fact that the capability of the RL agent depends on training time and state. In general, as more data is observed, the RL agent improves its behavior, ultimately outperforming the control prior in large portions of the domain. The core idea of our approach is to adapt the weighting between RL agent and control prior based on the agent's confidence.
As the RL agent improves over time, this induces a time-variant weighting mechanism. This time dependency leads to structural problems of prior formulations in uncertainty-adapted hybrid RL~\citep{cheng_control_2019, rana_bayesian_2023}.

We provide a unified view on hybrid RL that allows us to classify prior work within a general framework.
Analyzing this framework highlights the necessity for a novel adaptive hybrid RL formulation with descriptive, time-invariant dynamics.
We define the contextualized hybrid Markov decision process (MDP), introducing the adaptive weight as a context variable. Building upon this formulation, we propose the \emph{Contextualized Hybrid Ensemble Q-learning} (CHEQ) algorithm that systematically adapts the weighting between RL agent and control prior based on an uncertainty estimate of a critic ensemble. CHEQ combines the contextualized hybrid RL formulation with uncertainty-based weight adaption and existing ensemble-based acceleration techniques for data-efficient RL.

We evaluate our algorithm on a racing task~\citep{schier_learned_2023}, which requires operating a car close to its stability limits in order to achieve maximum return.
We find that compared to prior work in adaptive hybrid RL, the CHEQ algorithm shows (i) reduced failures during training, (ii) increased sample efficiency, and (iii) improved transfer behavior on unseen race tracks.

In summary, our main contributions are:
\begin{compactitem}
    \item A unified framework that allows us to classify existing approaches and reveal key limitations.
    \item A hybrid MDP formulation, introducing the adaptive weight as a context variable and thus addressing structural problems of prior work in hybrid RL with adaptive weighting.
    \item A novel hybrid RL algorithm, CHEQ, that systematically adapts the weighting between RL agent and control prior based on Q-ensemble disagreement.
\end{compactitem}
%
%
%
%
%
\section{Related Work}
\label{sec:related_work_hybrid_rl}
This section discusses relevant prior work combining RL and a control prior. We distinguish two types; hybrid RL with fixed and adaptive weighting between the RL agent and controller.
 
\fakepar{Hybrid Reinforcement Learning with Fixed Weighting}
Two concurrent works~\citep{silver_residual_2018, johannink_residual_2019} first combined RL and a control prior and introduced the term residual RL. In residual RL, the control prior is assumed to be fixed, and the RL agent learns a residual on top of this. In this work, we use the general term hybrid RL to include approaches that adapt the controller's weight.~\citet{silver_general_2018} and~\citet{johannink_residual_2019} show advantages of hybrid RL, such as sample efficiency, improved sim-to-real transfer, and robustness towards uncertainties. Hybrid RL with fixed weights has then successfully been applied to real robot insertion tasks~\citep{schoettler_deep_2020}, peg insertion under uncertainty~\citep{ranjbar_residual_2021}, driving~\citep{kerbel_residual_2022} and to learn a residual RL policy on top of a pre-trained RL agent~\citep{ceola_resprect_2024}. A fixed mixing, however, does not allow one to consider the improving capabilities of the RL agent.

\fakepar{Hybrid Reinforcement Learning with Adaptive Weighting}
\citet{daoudi_enhancing_2023} assume a given controller confidence function, employing a controller in instances of high confidence and an RL agent in other scenarios. Our work focuses on the RL agent's confidence and proposes to estimate the confidence based on a critic ensemble.
Similar to our approach, ~\citet{hoel_reinforcement_2020, hoel_tactical_2020} train an ensemble of bootstrapped Q-networks for a driving task with discrete actions. They evaluate the uncertainty as the coefficient of variation of Q-estimates and resort to safe fallback actions in case of high uncertainty. However, they do not combine controller and RL agent but switch between both. In this work, we investigate a seamless blending approach for continuous control. 
~\citet{rana_residual_2020} estimate the policy uncertainty using Monte-Carlo dropout and based on this uncertainty either sample from a residual policy or the controller alone. ~\citet{rana_multiplicative_2020} directly fuse a prior control distribution with an RL policy in a multiplicative fashion and anneal the influence of the control prior over training time.~\citet{rana_bayesian_2023} use a policy ensemble to estimate how certain the RL agent is in the current action. The combined action is then computed as the Bayesian posterior of control prior and policy distribution.
\citet{cheng_control_2019} use the TD-error as an uncertainty estimate and combine controller and RL agent based on this.
Both~\citet{rana_bayesian_2023} and~\citet{cheng_control_2019} base their adaption mechanism on a form of policy uncertainty. Both approaches train based on the combined action, which becomes brittle when facing large distributional shifts.
We further discuss this limitation in Section \ref{sec:context_hyb_rl}.
%
%
%
%
%
\section{Background}
\label{sec:background}
The following introduces the key components and the general concept of hybrid RL.

\fakepar{Reinforcement Learning} RL is a method for solving sequential decision problems based on the interaction between an agent and an environment ~\citep{sutton_reinforcement_2018}.
The environment is modeled as a discounted Markov decision process defined by the tuple $\mathcal{M} = (\mathcal{S}, \mathcal{A}, p, r, \rho_0, \gamma)$, with state space $\mathcal{S}$, action space $\mathcal{A}$, and start state distribution $\rho_0$. The commonly unknown transition function $p(\mathbf{s}_{t+1}, r_{t+1} \mid \mathbf{s}_t, \mathbf{a}_t^{\mathrm{RL}})$ describes transitions between states $\mathbf{s}_t \in \mathcal{S}$ and actions $\mathbf{a}^{\mathrm{RL}}_t \in \mathcal{A}$. During transitions, rewards $r_t \in \mathbb{R}$ are emitted according to a reward function $r_{t+1} \sim r(\mathbf{s}_t, \mathbf{a}^{\mathrm{RL}}_t)$. The objective of the RL agent is to learn a policy $\pi^{\mathrm{RL}}(\mathbf{a}^{\mathrm{RL}}_t \mid \mathbf{s}_t)$ that maximizes the expected cumulative sum of rewards discounted by $\gamma \in (0, 1)$. This results in the RL objective 
\begin{equation}
    J(\pi^\mathrm{RL}) = \max_{\pi^\mathrm{RL}} \mathbb{E}_{\pi^\mathrm{RL}, \mathcal{M}} \left[ \sum_{t=0}^{\infty} \gamma^t r_{t+1} \right].
    \label{eq:standard_rl_objective}
\end{equation}
The discounted sum of rewards is referred to as return and is accumulated along trajectories under the policy $\pi^\mathrm{RL}$ and the environment MDP $\mathcal{M}$.
State value functions condition expected return on a particular state $V^{\pi^\mathrm{RL}}(\mathbf{s}_t) = \mathbb{E}_{\pi^\mathrm{RL}, \mathcal{M}} \left[ \sum_{k=t}^{\infty} \gamma^{k-t} r_{k+1} \mid \mathbf{s}_t \right]$ while, action value or Q-functions condition expected return on specific state action pairs  $Q^{\pi^\mathrm{RL}}(\mathbf{s}_t, \mathbf{a}^{\mathrm{RL}}_t) = \mathbb{E}_{\pi^\mathrm{RL}, \mathcal{M}} \left[ \sum_{k=t}^{\infty} \gamma^{k-t} r_{k+1} \mid \mathbf{s}_t, \mathbf{a}^{\mathrm{RL}}_t \right]$.

\fakepar{Control Prior} The prior policy $\pi^{\mathrm{prior}}(\mathbf{a}^{\mathrm{prior}}_t \mid \mathbf{s}_t)$ represents prior knowledge for solving the RL objective \eqref{eq:standard_rl_objective}, while typically not providing the optimal policy over the whole domain $\mathcal{S} \times \mathcal{A}$. 
This work focuses on control priors based on classic control theory. These can be derived with limited effort in many applications and often provide a good baseline for interaction with $\mathcal{M}$. We assume a control prior that is time-invariant and without an internal state.

\fakepar{Hybrid Reinforcement Learning} Hybrid RL combines the control prior and the RL agent by blending their actions via some mixing function $\mathbf{a}^\mathrm{mix}_t =f(\mathbf{a}^\mathrm{prior}_t, \mathbf{a}^\mathrm{RL}_t, \boldsymbol{\lambda}_t)$ depending on a weight $\boldsymbol{\lambda}_t$. 
%
%
%
%
%
\section{A Unified View on Hybrid Reinforcement Learning} 
\label{sec:adaptiveproblem}
Next, we develop a unified view on hybrid RL that allows classifying prior methods (cf. Section \ref{sec:related_work_hybrid_rl}). In the standard RL setup depicted in Figure \ref{fig:standard_rl}, the RL agent $\pi^{\mathrm{RL}}$ interacts with the time-invariant MDP $\mathcal{M}$ with dynamics $p(\mathbf{s}_{t+1}, r_{t+1} \mid \mathbf{s}_t, \mathbf{a}^{\mathrm{RL}}_t)$ that represents the controlled system. Hybrid RL (see Figure \ref{fig:nocontex_hybrid_rl}) incorporates a control prior $\pi^{\mathrm{prior}}$ which requires reformulating the standard framework. Here, $\pi^{\mathrm{RL}}$ and $\pi^{\mathrm{prior}}$ apply a combined action $\mathbf{a}^{\mathrm{mix}}_t$ to $\mathcal{M}$. The mixing function $f$ generates a combined action by blending the individual actions based on a weighting vector $\boldsymbol{\lambda}_t$ provided by a weight adaption function $\Lambda$. Within this generalized framework, prior work in hybrid RL can be categorized based on the choice of $f$ and $\Lambda$.

\subsection{Mixing Function $f$} 
We consider mixing functions based on a weighted sum with a weighting vector $\boldsymbol{\lambda}_t = [\lambda^\mathrm{prior}_t, \lambda^\mathrm{RL}_t]^\top$
\begin{equation}
    \label{eq:mixing_function}
    \mathbf{a}^\mathrm{mix}_t = f(\mathbf{a}^\mathrm{prior}_t, \mathbf{a}^\mathrm{RL}_t, \boldsymbol{\lambda}_t) = \lambda^\mathrm{prior}_t \cdot \mathbf{a}^\mathrm{prior}_t + \lambda^\mathrm{RL}_t \cdot \mathbf{a}^\mathrm{RL}_t.
\end{equation}
This formulation allows to distinguish a \emph{residual} and a \emph{regularized} setting.

In the residual setting, $\lambda^\mathrm{prior}_t$ is typically constant while $\lambda^\mathrm{RL}_t$ can be variable $\boldsymbol{\lambda}_t = [1, \lambda^\mathrm{RL}_t]^\top$ \citep{silver_residual_2018, johannink_residual_2019, schoettler_deep_2020}. Thus, the RL agent interacts with the closed control loop between $\pi^{\mathrm{prior}}$ and $\mathcal{M}$ and learns a residual action on top of $\mathbf{a}^\mathrm{prior}_t$. Consequently, $\boldsymbol{\lambda}_t$ modulates the RL agent's impact on the closed loop dynamics. As the control prior is not scaled down, it might interpret the RL agent as a disturbance and counteract it~\citep{ranjbar_residual_2021}, which can limit the overall performance of residual formulations. 

In the regularized setting, both weights are adaptable such that $\lambda^\mathrm{prior}_t + \lambda^\mathrm{RL}_t = 1$~\citep{cheng_control_2019, rana_bayesian_2023}. This results in a mixing function of the form
\begin{equation}
    \mathbf{a}^\mathrm{mix}_t = f(\mathbf{a}^\mathrm{prior}_t, \mathbf{a}^\mathrm{RL}_t, \lambda^\mathrm{RL}_t) = (1-\lambda^\mathrm{RL}_t) \cdot \mathbf{a}^\mathrm{prior}_t + \lambda^\mathrm{RL}_t \cdot \mathbf{a}^\mathrm{RL}_t.
    \label{eq:apr-mixing-function}
\end{equation}
with $\lambda^\mathrm{RL}_t \in [0, 1]$.
In the limit $\lambda^\mathrm{RL}_t = 0$, the control prior interacts with $\mathcal{M}$ without the interference of the RL agent, while the regularized setting reduces to the standard RL problem for $\lambda^\mathrm{RL}_t = 1$. Thus, $\lambda^\mathrm{RL}_t$ indicates not only the impact of $\mathbf{a}^\mathrm{RL}_t$ but also whether the RL agent interacts with the open loop dynamics of $\mathcal{M}$ or the closed loop dynamics as in the residual setting. Consequently, the control prior can be interpreted as a regularization of the RL agent. Our proposed algorithm operates in the regularized setting, allowing the agent to take over complete control when $\lambda^\mathrm{RL}_t=1$.

\subsection{Weight Adaption Function $\Lambda$}
\label{subsec:weight_adaption_function}
Hybrid RL approaches can further be classified, based on the choice of the weight adaption function $\Lambda$ modulating the weighting vector $\boldsymbol{\lambda}_t$ of the mixing function $f$.

A large body of work, which we refer to as \emph{fixed-weight hybrid RL}~\citep{silver_residual_2018, johannink_residual_2019, schoettler_deep_2020, ranjbar_residual_2021, ceola_resprect_2024} chooses $\boldsymbol{\lambda}_t$ fixed throughout training. Neglecting the time- and state-dependent capabilities of the RL agent.

Approaches that adapt $\boldsymbol{\lambda}_t$, which we refer to as \emph{adaptive hybrid RL} methods, rely on different mechanisms. Scheduling approaches~\citep{rana_multiplicative_2020} change the weight explicitly with time, i.e. $\boldsymbol{\lambda}_t = \Lambda(t)$, typically increasing the weight of the RL agent as training progresses. Domain-based approaches~\citep{kulkarni_learning_2022,daoudi_enhancing_2023} adapt the weights based on the point of operation within the domain $\mathcal{S} \times \mathcal{A}$, i.e. $\boldsymbol{\lambda}_t = \Lambda(\mathbf{s}_t, \mathbf{a}_t)$.
Uncertainty-based approaches~\citep{cheng_control_2019, rana_bayesian_2023}
adapt the weight based on the confidence of the RL agent, indicated by an uncertainty estimate $u(\mathbf{s}_t, \mathbf{a}_t, t)$, giving more weight to the RL agent when it has high confidence.
Thus, they aim to leverage the benefits of the RL agent whenever possible, while resorting to a safe controller in situations where the RL agent has not seen enough data. The time dependency of the uncertainty estimate, however, increases the complexity of the hybrid RL setting, requiring a reformulation of the learning problem. In Section \ref{sec:context_hyb_rl}, we discuss the shortcomings of prior formulations and propose our own.
\begin{figure}[tb]
     \centering
     \begin{subfigure}[b]{0.32\columnwidth}
         \centering
        \includegraphics[width=\textwidth]{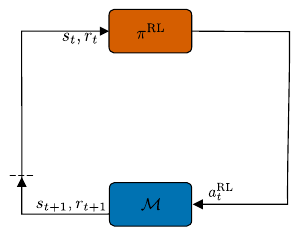}
         \caption{}
         \label{fig:standard_rl}
     \end{subfigure}
     \begin{subfigure}[b]{0.32\columnwidth}
         \centering
        \includegraphics[width=\textwidth]{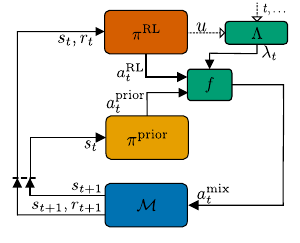}
         \caption{}
         \label{fig:nocontex_hybrid_rl}
     \end{subfigure}
     \begin{subfigure}[b]{0.32\columnwidth}
        \centering
        \includegraphics[width=\textwidth]{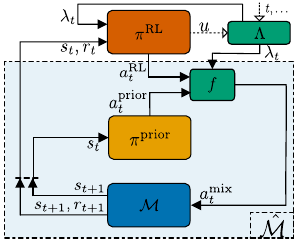}
         \caption{}
        \label{fig:contex_hybrid_rl}
    \end{subfigure}
     \caption{A standard RL setting (\subref{fig:standard_rl}), hybrid RL settings from prior work (\subref{fig:nocontex_hybrid_rl}) and our contextualized hybrid RL setting based on RL action $\mathbf{a}^\mathrm{RL}_t$ and weighting factor $\boldsymbol{\lambda}^\mathrm{RL}_t$ (\subref{fig:contex_hybrid_rl}).}
     \label{fig:rl_formulations}
\end{figure}
%
%
%
%
\section{Contextualized Hybrid Reinforcement Learning}
\label{sec:context_hyb_rl}
In Section \ref{subsec:context_hyb_rl}, we propose a novel contextualized formulation of the adaptive hybrid RL problem and illustrate its benefits over prior approaches in Section \ref{subsec:example_cartpole}. Based on that framework, we propose the Contextualized Hybrid Ensemble Q-learning (CHEQ) algorithm in Section \ref{subsec:cheq}.

\subsection{General Concept of Contextualized Hybrid Reinforcement Learning}
\label{subsec:context_hyb_rl}

Based on the unified view provided in Section \ref{sec:adaptiveproblem}, we propose a general formulation for the hybrid RL problem we call \emph{contextualized hybrid RL}. 

The environment in the hybrid setting not only consists of the controlled system $\mathcal{M}$ but also comprises the control prior, the mixing function, and the weight adaption function. We consider both the control prior and the mixing function to be time-invariant. In contrast, the weight adaption function $\Lambda$ can have time-varying behavior, i.e. $\boldsymbol{\lambda}_t = \Lambda( t, \dots)$, as discussed in Section \ref{subsec:weight_adaption_function}. This leads to time-varying dynamics of the hybrid environment, violating the assumption of time-invariance in the MDP formulation~\citep{bdr2023}. 
Instead, we exclude $\Lambda$ from the definition of the hybrid environment and introduce the adaptive weight vector $\boldsymbol{\lambda}_t$ as a context variable to the agent and the environment (see Figure \ref{fig:contex_hybrid_rl}).
We model the hybrid environment introducing the contextualized hybrid MDP $\hat{\mathcal{M}} = (\mathcal{S}, \mathcal{A}, \mathcal{W}, \hat{p}, r, \rho_0, \gamma)$ with $\mathcal{W}$ the set of weighting vectors $\boldsymbol{\lambda}_t$ and the contextualized dynamics function $\hat{p}(\mathbf{s}_{t+1}, r_{t+1} \mid \mathbf{s}_t, \mathbf{a}^{\mathrm{RL}}_t, \boldsymbol{\lambda}_t)$.

The MDP formulation $\hat{\mathcal{M}}$ induces the contextualized hybrid RL objective
\begin{equation}
    \hat{J}(\pi^\mathrm{RL}) = \max_{\pi^\mathrm{RL}} \mathbb{E}_{\pi^\mathrm{RL}, \hat{\mathcal{M}}} \left[ \sum_{t=0}^T \gamma^t r_{t+1}\right],
    \label{eq:hybrid_rl_objective}
\end{equation}
which enforces to learn a policy $\hat{\pi}^\mathrm{RL}(\mathbf{a}^{\mathrm{RL}}_t \mid \mathbf{s}^{\mathrm{RL}}_t, \boldsymbol{\lambda}_t)$ that maximizes expected return in $\hat{\mathcal{M}}$. Introducing $\boldsymbol{\lambda}_t$ as a context variable further yields the contextualized hybrid value functions
$\hat{V}^{\pi^\mathrm{RL}}(\mathbf{s}_t, \boldsymbol{\lambda}_t) = \mathbb{E}_{\pi^\mathrm{RL}, \hat{\mathcal{M}}} \left[ \sum_{k=t}^{\infty} \gamma^{k-t} r_{k+1} \mid \mathbf{s}_t, \boldsymbol{\lambda}_t\right]$ and 
$\hat{Q}^{\pi^\mathrm{RL}}(\mathbf{s}_t, \mathbf{a}^{\mathrm{RL}}_t, \boldsymbol{\lambda}_t) = \mathbb{E}_{\pi^\mathrm{RL}, \hat{\mathcal{M}}} \left[ \sum_{k=t}^{\infty} \gamma^{k-t} r_{k+1} \mid \mathbf{s}_t, \mathbf{a}^{\mathrm{RL}}_t, \boldsymbol{\lambda}_t \right]$. Thus, we can optimize \eqref{eq:hybrid_rl_objective} using standard RL methods by additionally conditioning on $\boldsymbol{\lambda}_t$.
The general mechanism of contextualized hybrid RL is illustrated in Algorithm \ref{alg:AH-RL}.

\begin{algorithm}
\caption{Contextualized Hybrid Reinforcement Learning }
\begin{algorithmic}[1]
\Require{RL policy $\hat{\pi}^\mathrm{RL}_\phi(\mathbf{a}^\mathrm{RL}_t \mid \mathbf{s}^\mathrm{RL}_t, \boldsymbol{\lambda}_t)$, control prior $\pi^\mathrm{prior}(\mathbf{a}^\mathrm{prior}_t \mid s)$, mixing function $f(\mathbf{a}^\mathrm{RL}_t, \mathbf{a}^\mathrm{prior}_t, \boldsymbol{\lambda}_t)$, weight adaption function $\Lambda$, replay buffer $\mathcal{D} \leftarrow \emptyset$.}
\For{each episode} 
\State Sample initial state $\mathbf{s}_0 \sim \rho_0$, initialize $\boldsymbol{\lambda}_0$.
\For{each step}
\State Sample RL action $\mathbf{a}^\mathrm{RL}_t \sim \pi^\mathrm{RL}_\phi\left(a^\mathrm{RL}_t \mid \mathbf{s}_t, \boldsymbol{\lambda}_t\right)$.
\State Sample control prior action $\mathbf{a}^\mathrm{prior}_t \sim \pi^\mathrm{prior}_\phi\left(a^\mathrm{RL}_t \mid \mathbf{s}_t\right)$.
\State Get combined action $\mathbf{a}^\mathrm{mix}_t= f(\mathbf{a}^\mathrm{RL}_t, \mathbf{a}^\mathrm{prior}_t, \boldsymbol{\lambda}_t)$. \label{alg:AH-RL-f}
\State Observe state transition $\mathbf{s}_{t+1}, r_{t+1} \sim p\left(\cdot, \cdot \mid \mathbf{s}_t,a^\mathrm{mix}_t\right)$.
\State Store $\left(\mathbf{s}_t, \mathbf{a}^\mathrm{RL}_t, \boldsymbol{\lambda}_t, \mathbf{s}_{t+1}, r_{t+1} \right)$ into replay buffer $\mathcal{D}$.
\State Get next adaptive weight $\boldsymbol{\lambda}_{t+1} = \Lambda $. \label{alg:AH-RL-lambda}
\State Sample set of transitions $\left(s, a, \lambda,  s^{\prime}, r\right) \sim \mathcal{D}$.
\State Optimize $\phi$ with respect to \eqref{eq:hybrid_rl_objective} using RL with sampled transitions.
\EndFor
\EndFor
\end{algorithmic}
\label{alg:AH-RL}
\end{algorithm}
Prior work takes different approaches to formulating the hybrid learning problem.
Approaches with a time-invariant weight adaption function, such as fixed-weight hybrid methods, include $\Lambda$ in the definition of the environment MDP $\bar{\mathcal{M}} = (\mathcal{S}, \mathcal{A}, \bar{p}, r, \rho_0, \gamma)$ with dynamics $\bar{p}(\mathbf{s}_{t+1}, r_{t+1} \mid \mathbf{s}_t, \mathbf{a}^{\mathrm{RL}}_t)$. This formulation is directly applicable to the standard RL objective \eqref{eq:standard_rl_objective}, however, does not generalize to time-varying adaption mechanisms such as uncertainty-adapted methods.
Approaches with a time-varying adaption mechanism~\citep{cheng_control_2019, rana_bayesian_2023} typically formulate the hybrid RL problem concerning the controlled system MDP and the combined action with dynamics $p(\mathbf{s}_{t+1}, r_{t+1} \mid \mathbf{s}_t, \mathbf{a}^{\mathrm{mix}}_t)$. This likewise yields a formulation that is directly applicable to \eqref{eq:standard_rl_objective}, however, this can lead to problems as the agent is unaware of the downstream mixing process. Furthermore, this introduces a distributional shift between trained policy and data-collecting behavior policy. The distributional shift can lead to training instability and divergence~\citep{kumar_conservative_2020,fujimoto_addressing_2018}.

\subsection{Illustrative Example}
\label{subsec:example_cartpole}
\begin{figure}[tb]
    \centering
    \begin{subfigure}[c]{0.28\textwidth}
        \centering
        \includegraphics[width=\textwidth]{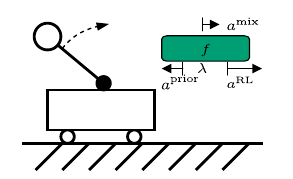}
        \caption{}
        \label{fig:cartpole_env}
    \end{subfigure}
    \begin{subfigure}[c]{0.35\textwidth}
        \centering
        \includegraphics[width=\textwidth]{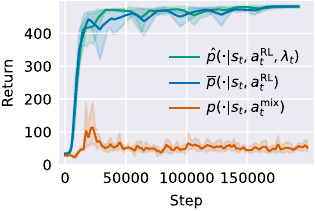}
        \caption{}
        \label{fig:cartpole_lambda_fix_return}
    \end{subfigure}
    \begin{subfigure}[c]{0.35\textwidth}
        \centering
        \includegraphics[width=\textwidth]{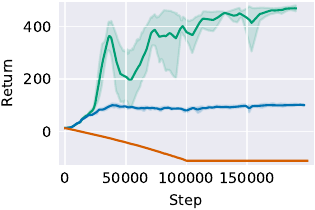}
        \caption{}
        \label{fig:cartpole_lambda_variable_return}
    \end{subfigure}
    \caption{We illustrate different hybrid RL formulations on a cart pole system (\subref{fig:cartpole_env}) with a biased control prior, pushing to the left. Return of hybrid agents with fixed $\lambda^\mathrm{RL}_t$ (\subref{fig:cartpole_lambda_fix_return})  and variable $\lambda^\mathrm{RL}_t$ (\subref{fig:cartpole_lambda_variable_return}). Only the contextualized hybrid RL agent observing dynamics $\hat{p}$ can cope with both scenarios.}
    \label{fig:cartpole_lambda_fix}
\end{figure}

We exemplify the strength of the contextualized hybrid RL formulation based on $\hat{\mathcal{M}}$ introduced in Section \ref{subsec:context_hyb_rl} by comparing it to prior approaches on the cart pole system depicted in Figure \ref{fig:cartpole_env}.
The goal is to balance the pole upright while keeping the cart close to its initial position. The system is controlled via continuous forces on the cart. We choose $\pi^\mathrm{prior}$ to apply a constant force to the left, which destabilizes formulations unconscious of the mixing process. We investigate a time-invariant fixed weight setting as well as a time-varying schedule-based weight adaption setting to highlight the capability of the respective formulations to deal with both scenarios.
This simple example illustrates that the contextualized hybrid MDP formulation can deal with destabilizing control priors and time-varying weight adaption functions while prior formulations fail.

\fakepar{Fixed Weighting}
First, we consider a residual setting with fixed weights $\lambda^{\mathrm{RL}}_t = \lambda^{\mathrm{prior}}_t = 0.5$. Figure \ref{fig:cartpole_lambda_fix_return} depicts the performance of RL agents trained under $\hat{\mathcal{M}}$, $\bar{\mathcal{M}}$, and $\mathcal{M}$ with respective dynamics $\hat{p}(\mathbf{s}_{t+1}, r_{t+1} \mid \mathbf{s}_t, \mathbf{a}^{\mathrm{RL}}_t, \boldsymbol{\lambda}_t)$, $\bar{p}(\mathbf{s}_{t+1}, r_{t+1} \mid \mathbf{s}_t, \mathbf{a}^{\mathrm{RL}}_t)$, and $p(\mathbf{s}_{t+1}, r_{t+1} \mid \mathbf{s}_t, \mathbf{a}^{\mathrm{mix}}_t)$. While agents trained under $\hat{\mathcal{M}}$ and $\bar{\mathcal{M}}$ learn to stabilize the cart pole, the hybrid formulation concerning $\mathcal{M}$ fails. 
When formulating the RL problem concerning $\mathbf{a}^{\mathrm{mix}}_t$ the RL agent observes the combined action in its data and therefore learns the combined action in its policy. This, however, neglects the fact that the policy action is mixed with the controller action before being applied to $\mathcal{M}$. Assuming the cart pole is not moving, and the pole is upright, an agent trained under $\mathcal{M}$ provides the optimal combined action, namely applying no force, while $\mathbf{a}^{\mathrm{prior}}_t$ pushes the pole to the left. This results in $\mathbf{a}^{\mathrm{mix}}_t$ pointing to the left, causing the pole to fall while giving the agent no mechanism to observe and counteract this phenomenon. Instead, formulating the hybrid RL problem concerning $\mathbf{a}^{\mathrm{RL}}_t$ allows the agent to observe the mixing mechanism and compensate for the destabilizing control prior.

\fakepar{Adaptive Weighting}
Second, we consider an adaptive hybrid RL problem with time-varying $\Lambda$. We choose a schedule-based approach with a regularizing mixing function \eqref{eq:apr-mixing-function} and $\lambda^{\mathrm{RL}}_t \in [0, 1]$ linearly increasing over time. Figure \ref{fig:cartpole_lambda_variable_return} shows the performance of agents trained under formulations based on $\hat{\mathcal{M}}$, $\bar{\mathcal{M}}$, and $\mathcal{M}$. While agents trained under $\bar{\mathcal{M}}$ and $\mathcal{M}$ fail, the formulation based on $\hat{\mathcal{M}}$ succeeds. In the beginning, when the RL agent is given only low weight, the formulation under $\mathcal{M}$ suffers from a high distributional shift between the action of the behavior policy $\mathbf{a}^{\mathrm{mix}}_t$ and the action of the target policy $\mathbf{a}^{\mathrm{RL}}_t$. The large distributional shift causes the agent to diverge~\citep{kumar_conservative_2020,fujimoto_addressing_2018}. Although the distributional shift decreases with increasing weight lambda, the agent does not manage to recover.
The formulation under $\bar{\mathcal{M}}$ fails due to the missing information about the time-variant behavior of the mixing process. The proposed contextualized hybrid RL formulation solves these issues by formulating the task concerning $\mathbf{a}^{\mathrm{RL}}_t$ and introducing the context variable $\boldsymbol{\lambda}_t$.

\subsection{Contextualized Hybrid Ensemble Q-learning (CHEQ)}
\label{subsec:cheq}
Based on the contextualized hybrid RL formulation introduced in Section \ref{subsec:context_hyb_rl}, we propose the Contextualized Hybrid Ensemble Q-learning (CHEQ) algorithm. At the heart of CHEQ is a critic ensemble that (i) provides an uncertainty estimate enabling an uncertainty-adapted hybrid RL mechanism, and (ii) allows to incorporate ensemble-based acceleration techniques for data-efficient RL. 

We base CHEQ on the Soft Actor-Critic (SAC)~\citep{haarnoja_soft_2018} algorithm and a regularizing mixing function \eqref{eq:apr-mixing-function} with $\boldsymbol{\lambda}_t = [(1-\lambda^\mathrm{RL}_t), \lambda^\mathrm{RL}_t]^\top$.
The weight adaption mechanism relies on a critic ensemble comprising of $E$ contextualized Q-functions with parameters $\theta_e$, $e \in \{1, \dots, E \}$ and corresponding target Q-functions with parameters $\bar{\theta}_e$, $e \in \{1, \dots, E \}$. We update the critics with the mechanism of Randomized Ensemble Double Q-learning (REDQ)~\citep{chen_randomized_2021} and enforce sufficient independence between Q-estimates using Bernoulli masking of the training data~\citep{osband_deep_2016, lee_sunrise_2021, mai_sample_2022}.
Model ensembles estimate parametric uncertainty, referred to as epistemic uncertainty, from disagreement between individual models within the ensemble. If different critics disagree about the outcome of taking action $\mathbf{a}^\mathrm{RL}_t$ in $\mathbf{s}_t$ while weighting with $\boldsymbol{\lambda}_t$, this indicates a weak understanding of the task in the particular area of $\mathcal{S} \times \mathcal{A} \times \mathcal{W}$. Thus, the control prior should be prioritized over the RL agent in such situations. Therefore, epistemic uncertainty represents a suitable quantity for adapting the weighting of control prior and RL agent.
We define epistemic uncertainty as the standard deviation of critic predictions
\begin{equation}
    u(\mathbf{s}_t,\mathbf{a}^\mathrm{RL}_t, \lambda^\mathrm{RL}_t) = \sqrt{\frac{1}{E} \sum_{e=1}^{E} \left( \hat{Q}_{\theta_e}(\mathbf{s}_t,\mathbf{a}^\mathrm{RL}_t, \lambda^\mathrm{RL}_t) - \mu (\mathbf{s}_t,\mathbf{a}^\mathrm{RL}_t, \lambda^\mathrm{RL}_t) \right)^2}
    \label{eq:ce_uncertainty}
\end{equation}
with $\mu(\mathbf{s}_t,\mathbf{a}^\mathrm{RL}_t, \lambda^\mathrm{RL}_t) =  \frac{1}{E} \sum_{e=1}^{E} \hat{Q}_{\theta_e}(\mathbf{s}_t,\mathbf{a}^\mathrm{RL}_t, \lambda^\mathrm{RL}_t)$.
We aim to give low weight to the RL agent in areas of high uncertainty and vice versa. 
Thus, the weight adaption function $\Lambda(u(\mathbf{s}_t, \mathbf{a}_t^\mathrm{RL}, \lambda^\mathrm{RL}_t))$ maps the critics epistemic uncertainty to the weighting factor $\lambda^\mathrm{RL}_t \in [\lambda_{\mathrm{min}}, \lambda_{\mathrm{max}}] \subseteq [0, 1]$ via the piece-wise linear function
\begin{equation}
     \lambda^{\mathrm{RL}}_{t+1} =
    \begin{cases}
        \lambda_{\mathrm{max}} & \text{if } u(\mathbf{s}_t, \mathbf{a}_t^\mathrm{RL}, \lambda^\mathrm{RL}_t) < u_{\mathrm{min}} \\
        \frac{ u(\mathbf{s}_t,\mathbf{a}^\mathrm{RL}_t, \lambda^\mathrm{RL}_t) - u_{\mathrm{max}} }{u_{\mathrm{min}} - u_{\mathrm{max}}} (\lambda_{\mathrm{ max}} - \lambda_{\mathrm{min}}) + \lambda_{\mathrm{min}} & \text{if } u(\mathbf{s}_t,\mathbf{a}^\mathrm{RL}_t, \lambda^\mathrm{RL}_t) \in[ u_{\mathrm{min}},  u_{\mathrm{max}}] \\
        \lambda_{\mathrm{min}}  & \text{if } u(\mathbf{s}_t,\mathbf{a}^\mathrm{RL}_t, \lambda^\mathrm{RL}_t) > u_{\mathrm{max}}.
    \end{cases}
\label{eq:ce_mixing_function}
\end{equation}
Besides providing an uncertainty estimate of the RL agent, the critic ensemble used in CHEQ has proven effective in mitigating overestimation bias ~\citep{thrun_issues_1993} in Q-learning-based approaches~\citep{lan_maxmin_2021, wang_adaptive_2021, chen_randomized_2021}. The Update-To-Data (UTD) ratio describes the number of gradient steps per environment interaction. Due to the reduction of the overestimation bias, the critic ensemble allows for increasing the UTD ratio while maintaining stable learning. This substantially improves the data efficiency of value-based actor-critic methods~\citep{chen_randomized_2021}. A detailed pseudocode algorithm of CHEQ is provided in Algorithm \ref{alg:CHEQ} of Appendix \ref{sec:app_algorithms_details}.

\section{Experiments}

We evaluate CHEQ on a racing task and compare it to standard RL, fixed-weighting hybrid RL, and state-of-the-art adaptive hybrid RL. 
In our experiments, CHEQ yields substantial improvements in (i) data efficiency compared to other hybrid methods, as well as (ii) exploration safety, and (iii) zero-shot transferability to unknown scenarios as compared to all competitor approaches.

\subsection{Experimental Setup}
We base our evaluation on a car racing setting adapted from~\citep{schier_learned_2023}. Achieving high returns requires advanced trajectory planning and control while operating the vehicle close to stability limits, including tire slip. The control prior is a trajectory-following task along the center line of the track using a Stanley controller~\citep{thrun_stanley_2006} for lateral and a proportional controller for longitudinal control. Further details are provided in Appendix \ref{sec:app_env_details}.

We compare CHEQ against the standard RL approaches SAC~\citep{haarnoja_soft_2018} and REDQ~\citep{chen_randomized_2021}, fixed-weighting hybrid RL based on SAC, and the state-of-the-art adaptive hybrid RL methods Controller Regularized RL (CORE)~\citep{cheng_control_2019} and Bayesian Controller Fusion (BCF)~\citep{rana_bayesian_2023}. In all experiments, we provide results for CHEQ with a high UTD ratio (CHEQ-UTD20) to demonstrate the capabilities of the approach and a low UTD ratio (CHEQ-UTD1) for a fair comparison to SAC-based methods.
All implementations\footnote{The code is available at \href{https://github.com/Data-Science-in-Mechanical-Engineering/cheq}{github.com/Data-Science-in-Mechanical-Engineering/cheq} .} are based on either the Clean RL library~\citep{huang_cleanrl_2022} or the original paper implementation~\citep{rana_bayesian_2023}. We provide a detailed description of the hyperparameter settings in Appendix \ref{sec:app_experimental_details_hyperparameter}.

We train all our approaches on ten random seeds and one fixed race track. We report return and cumulative training failures. Runs are considered a failure when the car leaves the track. For zero-shot transfer, we evaluate ten trained models per algorithm on ten unseen racetracks.
Return and failure plots show the respective mean (solid lines) and \qty{95}{\percent} confidence interval (shaded areas).

\subsection{Evaluation on the Car Racing Environment}
\label{sec:evaluation}

\begin{figure}[tb]
    \centering
    \begin{subfigure}[b]{0.32\textwidth}
        \centering
        \includegraphics[width=\textwidth]{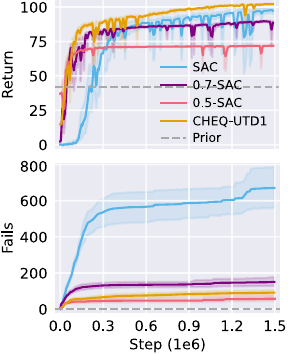}
        \caption{}
        \label{fig:cheq_vs_standard_rl_fixed}
    \end{subfigure}
    \begin{subfigure}[b]{0.32\textwidth}
        \centering
        \includegraphics[width=\textwidth]{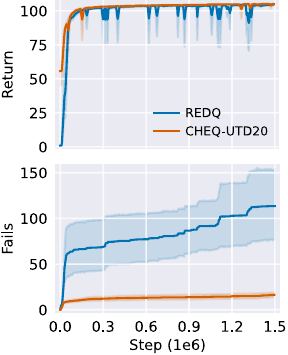}
        \caption{}
        \label{fig:cheq_vs_redq}
    \end{subfigure}
    \begin{subfigure}[b]{0.32\textwidth}
        \centering
        \includegraphics[width=\textwidth]{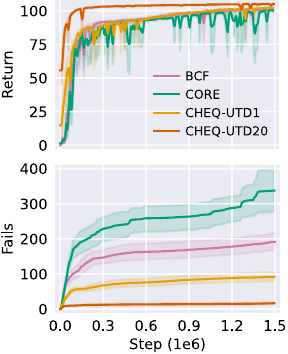}
        \caption{}
        \label{fig:cheq_vs_adaptive}
    \end{subfigure}
    \caption{Performance of all trained RL approaches in terms of evaluation return and training failures on the training track. Comparison of (\subref{fig:cheq_vs_standard_rl_fixed}) CHEQ with fixed-weight hybrid RL and standard RL, (\subref{fig:cheq_vs_redq}) increased UTD ratios and (\subref{fig:cheq_vs_adaptive}) prior work in adaptive hybrid RL.}
    \label{fig:training_performance}
\end{figure}

\begin{figure}[tb]
    \centering
    \begin{subfigure}[b]{0.32\textwidth}
        \centering
        \includegraphics[width=\textwidth]{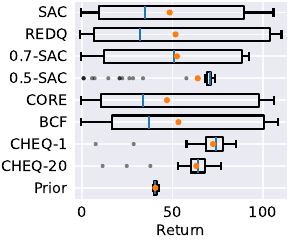}
        \caption{}
        \label{fig:transfer_boxplots}
    \end{subfigure}
    \begin{subfigure}[b]{0.32\textwidth}
        \centering
        \includegraphics[width=\textwidth]{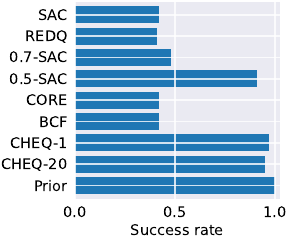}
        \caption{}
        \label{fig:transfer_success_rates}
    \end{subfigure}
    \begin{subfigure}[b]{0.32\textwidth}
        \centering
        \includegraphics[width=\textwidth]{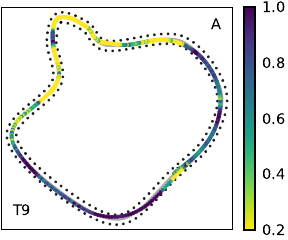}
        \caption{}
        \label{fig:uncertainty_map_transfer}
    \end{subfigure}
    \caption{Comparison of return (\subref{fig:transfer_boxplots}) and failures (\subref{fig:transfer_success_rates}) of ten trained models per algorithm on ten transfer tracks. Development of $\lambda^\mathrm{RL}$ of the UTD-20 agent for an exemplary transfer track (\subref{fig:uncertainty_map_transfer}).}
    \label{fig:zero_shot_transfer}
\end{figure}

We compare CHEQ to standard RL, fixed-weight hybrid RL, and adaptive hybrid RL concerning learning performance (see Figure \ref{fig:training_performance}) and zero-shot transfer to unknown tracks (see Figure \ref{fig:zero_shot_transfer}).

\fakepar{Comparison against Fixed-Weight Hybrid RL and Standard RL}
Comparing the CHEQ algorithm based on SAC (CHEQ-UTD1) to a standalone SAC agent and the control prior in Figure \ref{fig:cheq_vs_standard_rl_fixed} illustrates the general benefit of hybrid RL. While the control prior operates safely without failing, it shows limited performance due to the conservative driving policy. 
SAC shows strong asymptotic performance at the cost of frequent failures throughout training. CHEQ-UTD1 considerably outperforms SAC concerning data efficiency and exploration safety, learning faster with fewer failures while yielding comparable asymptotic performance. 
The comparison of CHEQ to fixed-weight hybrid RL methods further illustrates the advantage of an adaptive weighting scheme. The fixed-weight hybrid RL approaches (0.5-SAC, 0.7-SAC) combine the control prior with a SAC agent using the mixing function in \eqref{eq:apr-mixing-function} with $\lambda^{\mathrm{RL}}_t = 0.5$ and $\lambda^{\mathrm{RL}}_t = 0.7$, respectively. Here, the choice of $\lambda^{\mathrm{RL}}_t$ represents a trade-off between exploration safety and asymptotic performance, where a higher $\lambda^{\mathrm{RL}}_t$ enables better performance while reducing safety. A fixed weight of $\lambda^{\mathrm{RL}}_t = 0.5$ arguably reduces failures compared to CHEQ-UTD1, however, this comes at the cost of substantially lower performance.

\fakepar{Update-To-Data Ratio}
As discussed in Section \ref{subsec:cheq}, the critic ensemble of CHEQ allows the use of acceleration techniques originally proposed in the REDQ algorithm. Increasing the UTD ratio to \num{20} notably improves data efficiency as compared to SAC, both as a standalone RL algorithm (REDQ) and as an adaptive hybrid RL algorithm (CHEQ-UTD20). The speed-up in training helps to reduce training failures as REDQ reports a drastically reduced number of failures compared to SAC. The benefit is further amplified in the adaptive hybrid formulation of CHEQ-UTD20 as indicated by its strong initial performance and the ability to reduce the mean cumulative fails to less than \num{20}. 

\fakepar{Comparison against state-of-the-art Adaptive Hybrid RL}
Finally, Figure \ref{fig:cheq_vs_adaptive} compares CHEQ to the most relevant adaptive hybrid RL methods. CHEQ-UTD1 shows similar data efficiency and performance compared to CORE and BCF while considerably reducing accumulated fails. CHEQ-UTD20 substantially outperforms all competitor approaches in all performance metrics.
A more detailed hyperparameter analysis of the prior approaches, as well as results for reformulations of CORE and BCF as contextualized hybrid RL methods are provided in in Appendix \ref{sec:app_additional_results}.

\label{subsec:transfer}

\fakepar{Zero-shot Transfer}
Next, we perform a zero-shot transfer of the trained agents. Returns are depicted in Figure \ref{fig:transfer_boxplots} while Figure \ref{fig:transfer_success_rates} shows the success rate of the respective methods. CHEQ-UTD1, CHEQ-UTD20 and the control prior achieve a success rate of \qty{97}{\percent}, \qty{95}{\percent}, and \qty{100}{\percent}, respectively. The other standard and hybrid RL methods frequently fail in unseen scenarios. While the CHEQ variants fail slightly more often than the controller, they drive notably faster, i.e., they achieve higher returns. Figure \ref{fig:uncertainty_map_transfer} illustrates the adaption mechanism of CHEQ on one example track. In challenging and unseen curves, the agent gradually hands over to the control prior as can be seen in Figure \ref{fig:uncertainty_map_transfer}. We find that in the few failure cases (3 out of 100 for CHEQ-UTD1 and 5 out of 100 for CHEQ-UT20), the agent correctly identifies its uncertainty, and hands over to the controller, but the controller is unable to navigate the situation safely. We provide an illustration of all transfer tracks, as well as the weight adaption of CHEQ-UTD20 on these tracks in Appendix \ref{subsec:app_cheq}. In summary, CHEQ shows strong zero-shot transfer behavior, driving faster than the controller with only a few failure cases.

\begin{figure}[tb]
    \begin{subfigure}[b]{\textwidth}
        \centering
        \includegraphics[width=\textwidth]{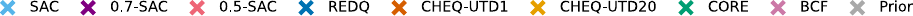}
        \label{fig:scatter_main_legend}
    \end{subfigure}
    \\
    \begin{subfigure}[b]{0.49\textwidth}
        \centering
        \includegraphics[width=\textwidth]{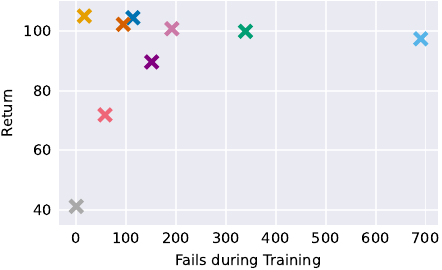}
        \caption{}
        \label{fig:scatter_main_training}
    \end{subfigure}
    \begin{subfigure}[b]{0.49\textwidth}
        \centering
        \includegraphics[width=\textwidth]{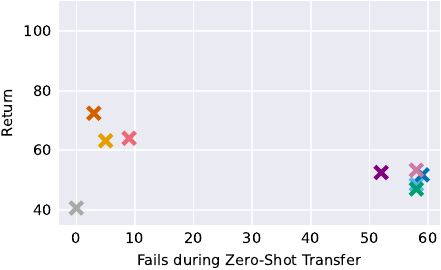}
        \caption{}
        \label{fig:scatter_main_transfer}
    \end{subfigure}
    \caption{Final return over number of fails during training (\subref{fig:scatter_main_training}) and zero-shot transfer (\subref{fig:scatter_main_transfer}).}
    \label{fig:scatter_main}
\end{figure}
\fakepar{Summary}
We summarize the trade-off between failures and asymptotic performance in Figure \ref{fig:scatter_main}. Figure~\ref{fig:scatter_main_training} illustrates the training results of the respective approaches while Figure~\ref{fig:scatter_main_transfer} depicts the transfer results.
Fixed weight hybrid RL effectively reduces failures as compared to standard SAC. This, however, comes at the cost of asymptotic performance.
Our adaptive CHEQ algorithm avoids this trade-off, achieving high return with only a few failures. In zero-shot transfer, the CHEQ agent again performs best due to its ability to detect unforeseen situations reliably and then fall back to the safe control prior.

\section{Conclusion}
\label{sec:conclusion}
This work addresses how to systematically combine an RL agent with a control prior.  We propose a novel formulation of the adaptive hybrid RL problem which introduces the adaptive weighting parameter as a context variable of the MDP, and based on this, propose the Contextualized Hybrid Ensemble Q-learning (CHEQ) algorithm. CHEQ combines a reliable critic uncertainty-based weight adaption mechanism with the data efficiency of critic ensemble methods, yielding substantially stronger results than state-of-the-art adaptive hybrid RL methods on a racing task concerning data efficiency, exploration safety, and transferability.

\subsubsection*{Acknowledgments}
\label{sec:ack}
We thank Paul Brunzema, Johanna Menn, and David Stenger for their helpful comments. We also thank Devdutt Subhasish and Lukas Kesper for their help with the cartpole example.
This work was funded in part by the German Federal Ministry of Education and Research (“Demonstrations- und Transfernetzwerk KI in der Produktion (ProKI-Netz)” initiative, grant number 02P22A010) and the German Federal Ministry for Economic Affairs and Climate Action (project EEMotion). Computations were performed with computing resources granted by RWTH Aachen University under the projects <thes1594>, <rwth1490>, and <rwth1501>.

\bibliography{main}
\bibliographystyle{rlc}

\clearpage
\appendix
\label{app:general}

\section{Algorithmic Details}
\label{sec:app_algorithms_details}
Algorithm \ref{alg:CHEQ} shows the pseudocode of the Contextualized Hybrid Ensemble Q-learning algorithm.

\begin{algorithm}[tbh]
\caption{CHEQ}
\label{alg:CHEQ}
    \begin{algorithmic}
\State Initialize control prior $\pi^\mathrm{prior}(\mathbf{a}^{\mathrm{prior}}_t \mid \mathbf{s}_t)$, contextualized RL policy $\hat{\pi}^\mathrm{RL}_\phi(\mathbf{a}^{\mathrm{RL}}_t \mid \mathbf{s}_t, \lambda^{\mathrm{RL}}_t)$, contextualized critic ensemble $\hat{Q}_{\theta_e}(\mathbf{s}_t, \mathbf{a}^{\mathrm{RL}}_t, \lambda^{\mathrm{RL}}_t)$, $e \in \{1, \dots, E \}$, contextualized target critic ensemble $\hat{Q}_{\bar{\theta}_e}(\mathbf{s}_t, \mathbf{a}^{\mathrm{RL}}_t, \lambda^{\mathrm{RL}}_t)$, $e \in \{1, \dots, E \}$, replay buffer $\mathcal{D} \leftarrow \emptyset$, weighting interval $[\lambda_{\mathrm{min}}, \lambda_{\mathrm{max}}]$, uncertainty limits $[u_{\mathrm{min}}, u_{\mathrm{max}}]$, UTD ratio $G$, Bernoulli masking rate $\kappa$, minimization targets $F$, Polyak averaging factor $\tau$
\For{each epoch}
    \State $\mathbf{s}_0 \sim \rho_0$, $\lambda^{\mathrm{RL}}_0 = \lambda_{\mathrm{min}}$
    \For{each epoch step}
        
        \State $\mathbf{a}^\mathrm{RL}_t \sim \hat{\pi}^\mathrm{RL}_\phi(\cdot \mid \mathbf{s}_t, \lambda^{\mathrm{RL}}_t)$ 
        \State $\mathbf{a}^\mathrm{prior}_t \sim \pi^\mathrm{prior}(\cdot \mid \mathbf{s}_t)$
        \State $\mathbf{a}^\mathrm{mix}_t = (1 - \lambda^{\mathrm{RL}}_t) \mathbf{a}^\mathrm{prior}_t + \lambda^{\mathrm{RL}}_t \mathbf{a}^\mathrm{RL}_t $
        \State $u(\mathbf{s}_t,\mathbf{a}^\mathrm{RL}_t, \lambda^{\mathrm{RL}}_t)$ according to \eqref{eq:ce_uncertainty}
            \State $\lambda_{t+1} = \Lambda(u(\mathbf{s}_t,\mathbf{a}^\mathrm{RL}_t, \lambda^{\mathrm{RL}}_t))$ according to \eqref{eq:ce_mixing_function}
        \State $\mathbf{s}_{t+1}, r_{t+1} \sim \hat{p}(\cdot, \cdot \mid \mathbf{s}_t,\mathbf{a}^\mathrm{RL}_t, \lambda^{\mathrm{RL}}_t)$
        \For{$e=1, \dots, E$}
            \State Sample Bernoulli Mask $m_t^e \sim Ber(\kappa)$
        \EndFor
        \State $\mathcal{D} \leftarrow \mathcal{D} \cup \{ (\mathbf{s}_t,\mathbf{a}^\mathrm{RL}_t, \lambda^{\mathrm{RL}}_t, \mathbf{s}_{t+1}, r_{t+1}), m_t^1, \dots, m_t^E\}$

        \For{$G$ updates}
            \State Sample mini-batch $\mathcal{B} = \{ (\mathbf{s}, \mathbf{a}^\mathrm{RL} , \lambda^{\mathrm{RL}}, \mathbf{s}^\prime, r \}$ from $\mathcal{D}$
            \State Sample a set $\mathcal{F}$ with $|\mathcal{F}| = F$ uniform at random from $\{ 1, \dots, E \}$
            \State $\Tilde{\mathbf{a}}^{\prime \mathrm{RL}} \sim \hat{\pi}^\mathrm{RL}_\phi( \cdot \mid \mathbf{s}^\prime, \lambda^{\mathrm{RL}}) $
            \State $y = r + \gamma \left( \min_{e \in \mathcal{F}} \hat{Q}_{\bar{\theta}_e}(\mathbf{s}^\prime, \Tilde{\mathbf{a}}^{\prime \mathrm{RL}}, \lambda^{\mathrm{RL}}) - \alpha \log \hat{\pi}^\mathrm{RL}_\phi( \Tilde{\mathbf{a}}^{\prime \mathrm{RL}} \mid \mathbf{s}^\prime, \lambda^{\mathrm{RL}}) \right)$
            \For{$e = 1, \dots, E$}
                \State Update $\theta_e$ with gradient descent using 
                \State $\mathbbm{1}_{m^e} \nabla_{\theta_e} \frac{1}{|\mathcal{B}|} \sum_{(\mathbf{s}, \mathbf{a}^\mathrm{RL}, \lambda^{\mathrm{RL}}, r, \mathbf{s}^\prime) \in \mathcal{B}} \left( \hat{Q}_{\theta_e}(\mathbf{s}, \mathbf{a}, \lambda^{\mathrm{RL}}) -y \right)^2$
                \State $\bar{\theta}_e \leftarrow \tau \bar{\theta}_e + (1 - \tau) \theta_e $
            \EndFor
        \EndFor
        \State update $\phi$ with gradient ascent using
        \State $\Tilde{\mathbf{a}}_{\mathrm{RL}} \sim \hat{\pi}^\mathrm{RL}_\phi( \cdot \mid \mathbf{s}, \lambda^{\mathrm{RL}}) )  $
        \State $\nabla_\phi \frac{1}{|\mathcal{B}|} \sum_{\mathbf{s} \in \mathcal{B}} \left( \frac{1}{E} \sum_{e=1}^E \hat{Q}_{\theta_e}(\mathbf{s},\Tilde{\mathbf{a}}_{\mathrm{RL}}, \lambda^{\mathrm{RL}}) - \alpha \log \hat{\pi}^\mathrm{RL}_\phi( \Tilde{\mathbf{a}}_{\mathrm{RL}} \mid \mathbf{s}, \lambda^{\mathrm{RL}}) ) \right)$
    \EndFor
\EndFor
    \end{algorithmic}
\end{algorithm}
%
\subsection{Hyperparameters Settings}
\label{sec:app_experimental_details_hyperparameter}
We build our SAC implementation based on CleanRL~\citep{huang_cleanrl_2022}.
All SAC-specific hyperparameters are kept consistent between all approaches and reported in Table \ref{tab:SAC_racing_hyperparameters}.

\begin{table}[htbp]
    \centering
    \caption{Shared Hyperparameters.}
    \begin{tabular}{| l | l |}
        \hline
        Hyperparameter &  Value\\
        \hline
        number of steps & $\num{1.5e6}$ \\
        batch size & 256 \\
        learning rate actor & \num{3e-4} \\
        learning rate critic & \num{3e-4} \\
        target entropy $H_t$ & \num{-3} \\
        replay buffer size & \num{1e6} \\
        discount factor $\gamma$ & \num{0.99} \\
        gradient update start & \num{1e3} steps \\
        Polyak averaging factor $\tau$ & \num{0.005} \\     
        \hline
    \end{tabular}
    \label{tab:SAC_racing_hyperparameters}
\end{table}

CHEQ (UTD1 and UTD20) uses an ensemble of $E = 5$ critics. We set the upper bound of the uncertainty as $u_\mathrm{max} = 0.15$ and the lower bound as $u_\mathrm{min} = 0.03$. Further we set $\lambda_\mathrm{max} = 1.0$ and $\lambda_{min} = 0.2$. We use a Bernoulli masking rate of $\kappa = 0.8$ and $F=2$ minimization targets.

BCF trains an ensemble of policy networks. We maintain the original ensemble size from~\citep{rana_bayesian_2023} which uses ten policy networks. We set the standard deviation of the control prior in BCF as $\sigma^\mathrm{prior} = 6.0$.

For the uncertainty estimate in CORE, we set $A=7$, $C=0.02$. Note that in the original paper, $A$ is denoted as $\lambda_\mathrm{max}$, which we change to avoid ambiguous notation.

SAC uses a UTD ratio of \num{1}. REDQ implementation uses an ensemble size of \num{5} and a UTD ratio of \num{20}.

For all algorithms, we include a random sampling phase for the first \num{1e3} steps where we sample the RL action uniformly random and do not update our agent. In this setting, we keep $\lambda^\mathrm{RL}_t$ small for the hybrid agents. For CHEQ we vary $\lambda^\mathrm{RL}$ between [0.2, 0.3]. As CORE and BCF are unable to observe changes in $\lambda^\mathrm{RL}$ we use a fixed $\lambda^\mathrm{RL}=0.2$ which has shown to be favorable in our experiments. After the random sampling phase, agent training starts, but $\lambda^\mathrm{RL}$ is kept small for another \num{4e3} steps and afterward, $\lambda$ adaption starts. 

For performance evaluation, we conduct a greedy evaluation run every \num{20} episodes. Evaluation happens in an adapted setting, together with the controller where the weight is calculated as in the training procedure.

\section{Environment Details}
\label{sec:app_env_details}

\subsection{Racing Environment}
\label{sec:app_env_details_racing}
\begin{figure}[htbp] 
    \centering
    \begin{subfigure}{0.53\textwidth}
        \includegraphics[width=\textwidth]{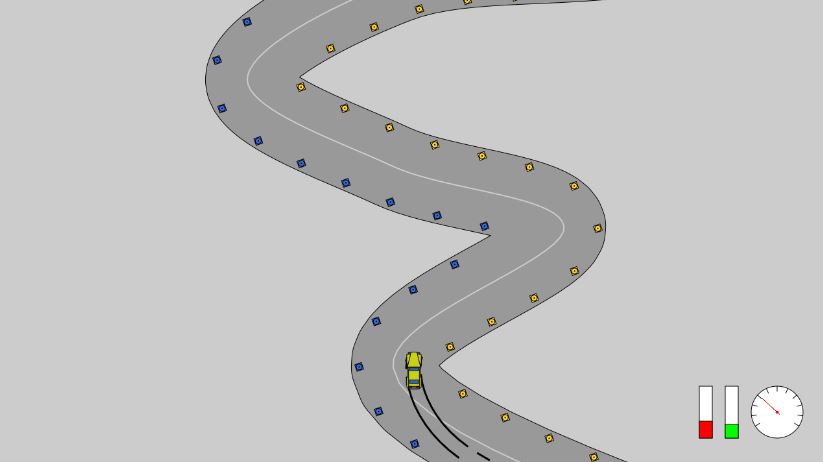} 
        \caption{}
    \end{subfigure}
    \begin{subfigure}{0.45\textwidth}
        \includegraphics[width=\textwidth]{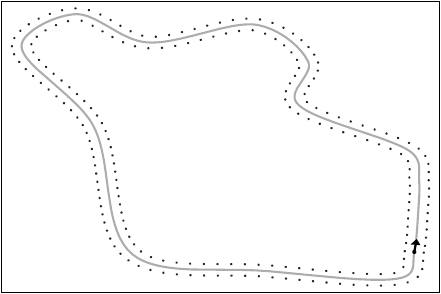} 
        \caption{}
    \end{subfigure}
    \caption{Racing task and the training track.}
    \label{fig:racing_scenario}
\end{figure}

We test our agent on the simulated racing task adapted from \citet{schier_learned_2023}. Figure \ref{fig:racing_scenario} shows an example of the environment.

The vehicle uses a dynamic single-track model with a coupled Dugoff tire model. The throttle, brake, and steering are continuous actions. The vehicle is a front-wheel drive. The RL agent may learn to control brake balance by applying throttle and brake individually.
We define the state of the RL agent as $\mathbf{s}_t = (v_x, v_y, \omega, \beta, o_\mathrm{track})$, with the ego vehicle's velocity vector $\mathbf{v}_\mathrm{ego} = (v_x, v_y)$ in vehicle reference frame, steering angle $\beta$, and yaw rate $\omega$.
The observation of the track $o_\mathrm{track} = (\mathbf{x},\mathbf{y})^T$ is given as a vector of \num{20} Cartesian distances $(x_i,y_i)$ to the centerline of the track. The points are sampled equidistantly from the \qty{60}{\metre} track segment ahead. 

We use the original reward formulation from \citet{schier_learned_2023} where the RL agent receives a penalty $r_\mathrm{collision}$ whenever it collides with the track boundary and a penalty $r_\mathrm{fail}$ for leaving the track with the center of mass. The latter case also terminates the episode. The RL agent receives a positive reward for driving fast: the scalar projection of its velocity vector $v_\mathrm{ego}$ onto the forward track direction $n_\mathrm{track}$. The complete reward is then given by 
\begin{equation}
    r(s, a)=-r_{\mathrm{fail}}-0.2 \cdot r_{\mathrm{collision}}+0.01 \cdot \mathbf{n}_{\mathrm{track}} \cdot \mathbf{v}_{\mathrm{ego}}.
\end{equation}

\subsection{Control Prior}
\label{sec:app_env_details_controller}
For the racing task, we design a simple path-following controller with adaptive speeds. For the lateral control, we use a Stanley Controller~\citep{thrun_stanley_2006} following the steering control law
\begin{equation*}
    \delta(t) = \psi(t) + \frac{k_\mathrm{cross}\cdot e(t)}{v(t) + k_\mathrm{soft}},
\end{equation*}
where $\psi(t)$ denotes the heading error, $e(t)$ denotes the crosstrack error of the front axle and $v(t)$ describes the velocity of the vehicle.

For the longitudinal control, we design two symmetric P-controllers; one for the brake and one for the throttle. First, we compute the target velocity dependent on the curve radius $R(t)$ of the track directly in front of the vehicle as
\begin{equation*}
    v_\mathrm{target}(t) = \min\{k_r \cdot R(t), v_\mathrm{max}\},
\end{equation*}
where $v_\mathrm{max}$ is the maximum desired velocity. Then, we design the throttle control as
\begin{equation*}
    \mathrm{throt}(t) = \begin{cases}k_v(t)*(v_{target}(t) - v(t)), &\quad v_{target}(t) - v(t) \geq 0\\
0, &\quad \text{else}\end{cases}
\end{equation*}
and the brake control as
\begin{equation*}
    \mathrm{br}(t) = \begin{cases}k_v(t)*(v(t) - v_{target}(t)), &\quad v_{target}(t) - v(t) \leq 0\\
0, &\quad \text{else,}\end{cases}
\end{equation*}
with shared gain $k_v$. Following this control law, the control prior accelerates if it is going too slow and brakes if it is going too fast. It never uses the brake and the throttle at the same time.

To avoid aggressive braking behavior when the RL agent hands over to the controller in risky situations (high velocity around curves), we additionally introduce a simple clipped linear gain schedule on $k_v$ attenuating the braking control for higher velocities as
\begin{equation*}
    k_v(t) = \mathrm{clip}\left(\frac{k_v^\mathrm{max} - k_v^\mathrm{min}}{v_\mathrm{low} - v_\mathrm{high}}(v(t) - v_\mathrm{low}) + k^\mathrm{max}_v; k_v^\mathrm{max}, k_v^\mathrm{min}\right).
\end{equation*}

We tuned the controller gains and coefficients to $k_\mathrm{cross} = 0.5[\si{1 \per\second}]$, $k_\mathrm{soft}= 1[\si{\metre \per\second}]$, $k_r=0.4[\si{1 \per\second}]$, $v_\mathrm{max}=8[\si{\metre \per\second}]$, $k^\mathrm{max}_v=0.25[\si{\second \per\metre}]$, $k_v^\mathrm{min}=0.05[\si{\second \per\metre}]$, $v_\mathrm{low}=8[\si{\metre \per\second}]$ and $v_\mathrm{high}=28[\si{\metre \per\second}]$.

\section{Additional Results and Ablation Study}
\label{sec:app_additional_results}
In this section, we formulate contextualized hybrid variants of CORE and BCF, C-CORE and C-BCF.
Further, we present additional results on hyperparameter sensitivity and the distribution of $\lambda^\mathrm{RL}$ for CHEQ-UTD1, CORE, and BCF.  For a reasonable comparison, we focus mainly on CHEQ-UTD1 using a UTD ratio of \num{1}.

\subsection{Contextualized Hybrid Variants of Prior Work}
\label{sec:app_c_bcf_core}
To further substantiate our claim that the contextualized hybrid RL formulation aids the training progress, we developed contextualized variants of the CORE and BCF algorithm, which we call C-CORE and C-BCF. To use the adaptive weight as a context variable, a weighting parameter $\lambda_t^\mathrm{RL}$ needs to be determined.

The CORE algorithm comes with a direct weight estimate\citep{cheng_control_2019}, which can be written as
\begin{equation*}
    \lambda_t^\mathrm{RL} = \frac{1}{1 + \lambda_t^\mathrm{CORE}}
\end{equation*}
where $\lambda_t^\mathrm{CORE}=A(1-e^{-C\abs{\delta_{t-1}}})$ with the TD-error $\delta_{t-1}$ and $C$, $A$ being tuning parameters\footnote{CORE~\citep{cheng_control_2019} uses the term $\lambda_\mathrm{max}$ instead of $A$. As we use $\lambda_\mathrm{max}$ in a different context, we stick to $A$ here.}.

For C-BCF, we derive a pseudo-weight, as the BCF algorithm~\citep{rana_bayesian_2023} does not have a straightforward weighting factor $\lambda^\mathrm{RL}$. In BCF, at timestep $t-1$, the fusion of the prior $\mathcal{N}_{\psi, t-1}(\mu_{\psi, t-1}, \sigma_{\psi, t-1})$ and the SAC-ensemble $\mathcal{N}_{\pi, t-1}(\mu_{\pi, t-1}, \sigma_{\pi, t-1})$ results in a Gaussian distribution with mean
\begin{equation*}
    \mu_{\mathrm{fuse}, t-1} = \frac{\sigma_{\psi,t-1}^2}{\sigma_{\psi,t-1}^2 + \sigma_{\pi,t-1}^2} \cdot \mu_{\pi, t-1} + 
    \frac{\sigma_{\pi,t-1}^2}{\sigma_{\psi,t-1}^2 + \sigma_{\pi, t-1}^2} \cdot \mu_{\psi, t-1}.
\end{equation*}

Thus, in BCF the weight has the dimension of the action space, whereas the contextualized mechanism requires a scalar weight $\lambda^{\mathrm{RL}}_t$.
For C-BCF, we compute a scalar weight
\begin{equation*}
    \lambda^\mathrm{RL}_t = \frac{1}{N}\sum_{i=1}^N \left(\frac{\sigma_{\psi, t-1}^2}{\sigma_{\psi, t-1}^2 + \sigma_{\pi,t-1}^2}\right)_i.
\end{equation*}
The next action $\mathbf{a}^\mathrm{mix}_t$ is computed according to Equation \ref{eq:apr-mixing-function} where
$\mathbf{a}^\mathrm{RL}_t \sim \mathcal{N}_{\pi, t}$ and $\mathbf{a}^\mathrm{prior}_t = \mu_{\psi, t}$.

For C-BCF and C-CORE we use the same warm-up phase as for our CHEQ agent.

\subsection{Additional Results and Hyperparameter Tuning for CHEQ}
\label{subsec:app_cheq}
\begin{figure}[htbp]
    \centering
    \begin{subfigure}[b]{0.49\textwidth}
        \centering
        \includegraphics[width=\textwidth]{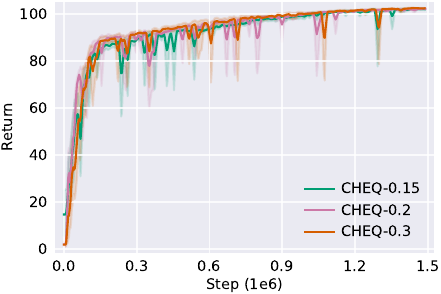}
        \caption{}
        \label{fig:evaluation-reward-comparison-CHEQ}
    \end{subfigure}
    \hfill
    \begin{subfigure}[b]{0.49\textwidth}
        \centering
        \includegraphics[width=\textwidth]{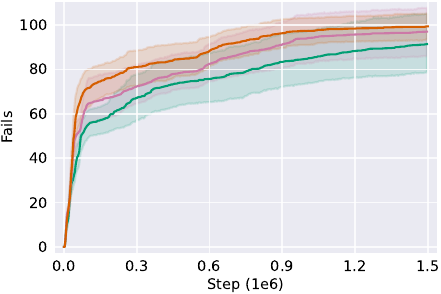}
        \caption{}
        \label{fig:fails-comparison-CHEQ}
    \end{subfigure}
    \caption{Comparison of CHEQ-UTD1 with different $u_\mathrm{max}$ thresholds. Plotting return (\subref{fig:evaluation-reward-comparison-CHEQ}) and number of fails (\subref{fig:fails-comparison-CHEQ}) for the racing environment.}
    \label{fig:comparisons_apr_ce}
\end{figure}
\begin{figure}[tb]
    \centering
    \includegraphics[width=\textwidth]{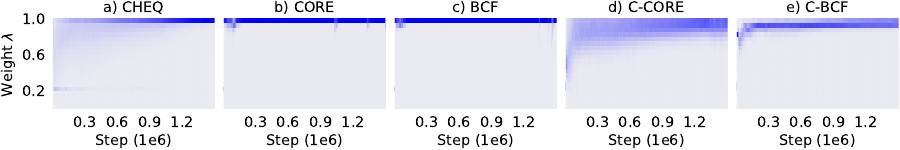}
    \caption{Distributions of $\lambda^\mathrm{RL}$ over training steps shown for all hybrid agents.}
    \label{fig:racing_lambda_distribution}
\end{figure}
\begin{figure}[htbp]
    \centering
    \includegraphics[width=\textwidth]{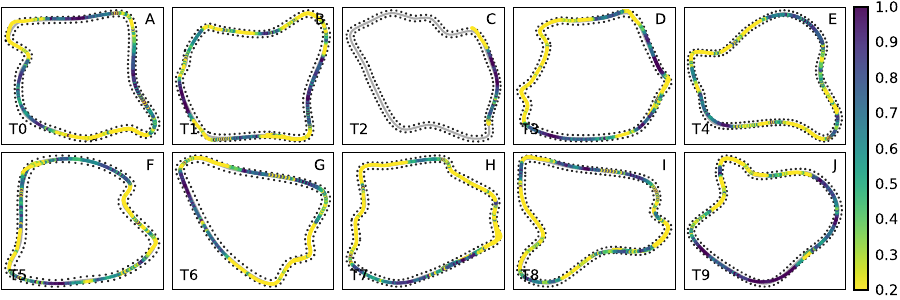}
    \caption{The 10 transfer tracks and the corresponding zero-shot transfer of one seed of CHEQ-UTD20.}
    \label{fig:transfer_track_plots}
\end{figure}
Our algorithm has only two important hyperparameters, upper and lower bounds of the uncertainty $u_\mathrm{max}$ and $u_\mathrm{min}$. We chose these hyperparameters by conducting one training run and investigating the uncertainty range within this run. CHEQ is generally robust against changes in these thresholds. We observe slightly lower final return and fewer fails for lower upper bounds $u_\mathrm{max}$. This is to be expected as frequent handover to the control prior results in lower velocities and thus lower return. Figure \ref{fig:comparisons_apr_ce} shows the return and the number of fails during training for our CHEQ-UTD1 variants. For CHEQ-UTD20 we were able to use the same upper and lower uncertainty bounds as for CHEQ-UTD1.

Figure \ref{fig:racing_lambda_distribution} shows the distribution of the weight $\lambda^\mathrm{RL}$ over the training progress. We find that for CHEQ-UTD1 the agent starts with an almost uniform distribution of the weight and slowly moves towards a $\lambda^\mathrm{RL}=1$ regime. Even in later training stages, the agent hands over to the controller from time to time. 

Lastly, we investigated the transfer behavior of the CHEQ-UTD20 agent further. Figure \ref{fig:transfer_track_plots} shows the ten transfer tracks. We plot $\lambda^\mathrm{RL}$ over the track for one evaluated model. Here, we find that the agent frequently becomes uncertain and hands over to the controller, especially in unknown curves. In plot C we see one of the two failure cases (out of 100 runs) that we experience during transfer. We find that the agent becomes uncertain and hands over to the controller. In this specific scenario, however, the controller is not able to safely navigate the situation and leaves the track. 

\subsection{Additional Results and Hyperparameter Tuning for CORE and C-CORE}
\begin{figure}[htbp]
    \centering
    \begin{subfigure}[b]{0.49\textwidth}
        \centering
        \includegraphics[width=\textwidth]{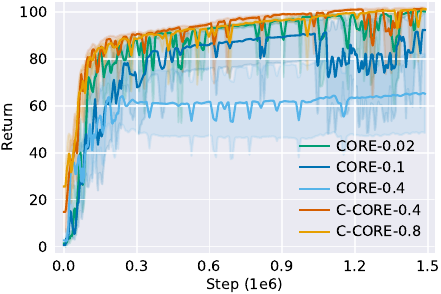}
        \caption{}
        \label{fig:reward_comparison_cores}
    \end{subfigure}
    \hfill
    \begin{subfigure}[b]{0.49\textwidth}
        \centering
        \includegraphics[width=\textwidth]{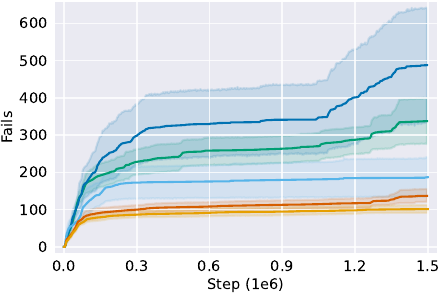}
        \caption{}
        \label{fig:fails_comparison_cores}
    \end{subfigure}
    \caption{Comparison of return (\subref{fig:reward_comparison_cores}) and number of fails (\subref{fig:fails_comparison_cores}) CORE and C-CORE runs with different $C$ parameters.}
    \label{fig:comparison_cores}
\end{figure}
In Figure \ref{fig:comparison_cores} we compare different parameters $C$. We find stable training progress and high return for $C=0.02$ but since this uses high $\lambda^\mathrm{RL}$ values, this setting results in many failures. Figure \ref{fig:racing_lambda_distribution} illustrates the high $\lambda$ regime of the $C=0.02$ agent.
C-CORE using our contextualized hybrid framework, notably outperforms CORE in terms of asymptotic return, training stability, and the number of training failures.
Using the contextualized formulation, C-CORE can use a much wider $\lambda^\mathrm{RL}$  distribution (see Figure \ref{fig:racing_lambda_distribution}).

\subsection{Additional Results and Hyperparameter Tuning for BCF and C-BCF}
\begin{figure}[htbp]
    \centering
    \begin{subfigure}[b]{0.49\textwidth}
        \centering
        \includegraphics[width=\textwidth]{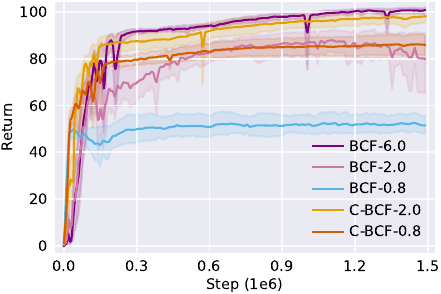}
        \caption{}
        \label{fig:reward_comparison_bcfs}
    \end{subfigure}
    \hfill
    \begin{subfigure}[b]{0.49\textwidth}
        \centering
        \includegraphics[width=\textwidth]{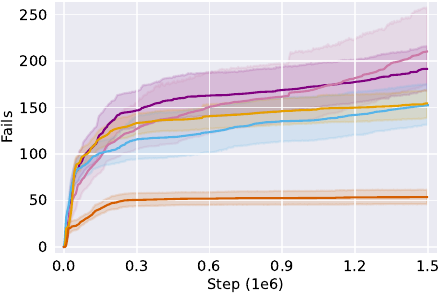}
        \caption{}
        \label{fig:fails_comparison_bcfs}
    \end{subfigure}
    \caption{Comparison of return (\subref{fig:reward_comparison_bcfs}) and number of fails (\subref{fig:fails_comparison_bcfs}) of BCF and C-BCF runs with different $\sigma_{prior}$ parameters.}
    \label{fig:comparison_bcfs}
\end{figure}

The BCF algorithm is sensitive to the parameters of the uncertainty threshold, which in this case is the variance of the control prior $\sigma^\mathrm{prior}$. Higher variances, lead to less weight on the control prior and thus high $\lambda^\mathrm{RL}$ regimes. In Figure \ref{fig:comparison_bcfs} we compared different parameters $\sigma^\mathrm{prior}$. We find stable training progress and high return for $\sigma^\mathrm{prior}=6$. However, this setting, as expected, uses a $\lambda^\mathrm{RL}$ regime close to one and thus results in a high number of failures. Figure \ref{fig:racing_lambda_distribution} illustrates this regime.

Our C-BCF variant can resolve this problem only partially. Due to its construction, the BCF algorithm has a separate weighting factor for each action of which we take the mean. In addition, our pseudo $\lambda^\mathrm{RL}$ factor is only a rough estimate of the actual mixing as BCF samples from the posterior distribution. Both factors result in information loss and make the weight $\lambda^\mathrm{RL}$ only a rough estimate. We find that C-BCF-2.0 and BCF-6.0 achieve similar asymptotic performance, while C-BCF-2.0 leads to fewer failures.

\subsection{Return vs. Failure Comparison for all trained Models}
\begin{figure}[htbp]
    \centering
    \includegraphics[width=\textwidth]{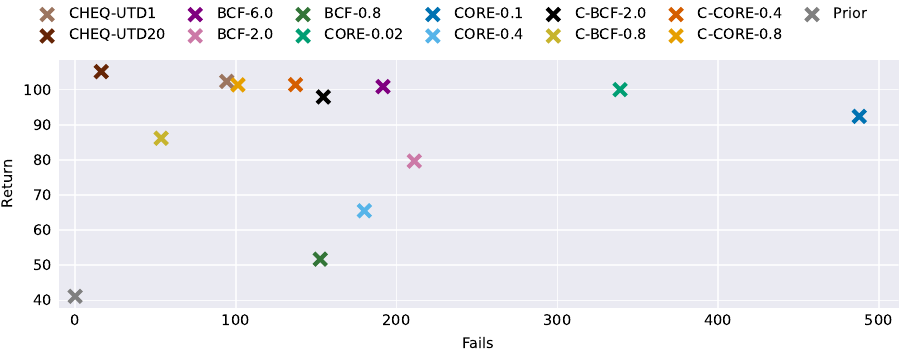}
    \caption{Scatter plot of the return and number of fails for different hyperparams for BCF, CORE, C-BCF, C-CORE. CHEQ-UTD20, CHEQ-UTD1 + Prior in Comparison}
    \label{fig:scatter_appendix}
\end{figure}

Figure \ref{fig:scatter_appendix} shows a scatter plot of the final return and the accumulated failures during training for all hybrid algorithms discussed in this paper. This final comparison shows that if prior methods are trained with the contextualized framework and tuned well ( C-BCF-0.8, C-CORE-0.4, C-CORE-0.8), they achieve high returns while maintaining fewer failures than their non-contextualized counterparts. Our algorithm (CHEQ-UTD1, CHEQ-UTD20) achieve the highest return while maintaining the lowest number of cumulative failures.

\end{document}